%% file: root.tex
\documentclass{article}

\usepackage[nonatbib,preprint]{neurips_2022}

\usepackage[utf8]{inputenc} %
\usepackage[T1]{fontenc}    %
\usepackage{hyperref}       %
\usepackage{url}            %
\usepackage{booktabs}       %
\usepackage{amsfonts}       %
\usepackage{nicefrac}       %
\usepackage{microtype}      %
\usepackage{xcolor}         %

\usepackage{tabularx}       %
\usepackage{graphicx}       %
\usepackage{mathtools} %
\usepackage{multirow}
\usepackage{amsmath}
\usepackage{amsthm}
\usepackage{amssymb}%
\usepackage{pifont}%
\usepackage{subcaption}
\usepackage{lscape}

\usepackage[ruled,noend]{algorithm2e}
\usepackage{wrapfig}

\newtheorem{nutheorem}{Theorem}
\DeclareMathOperator*{\argmax}{arg\,max}
\DeclareMathOperator*{\argmin}{arg\,min}
\DeclareMathOperator*{\KL}{KL}
\DeclareMathOperator*{\softmax}{softmax}
\DeclareMathOperator*{\entropy}{H}
\DeclareMathOperator*{\Kth}{\textit{K}-th}
\DeclareMathOperator\supp{supp}

\title{Know Your Boundaries: The Necessity of \\ Explicit Behavioral Cloning in Offline RL}

\author{%
  Wonjoon Goo and Scott Niekum\\
  Department of Computer Science\\
  The University of Texas at Austin\\
  \texttt{\{wonjoon,sniekum\}@cs.utexas.edu} \\
}

\begin{document}

\input{_main}

\bibliographystyle{plain}
{\small
\bibliography{root}}

\newpage
\appendix

\setcounter{figure}{0} \renewcommand{\thefigure}{A.\arabic{figure}}
\setcounter{table}{0} \renewcommand{\thetable}{A.\arabic{table}}
\setcounter{footnote}{0}

\input{_appendix}

\end{document}

%% file: _main.tex
\maketitle

\begin{abstract}
  We introduce an offline reinforcement learning (RL) algorithm that explicitly clones a behavior policy to constrain value learning.
  In offline RL, it is often important to prevent a policy from selecting unobserved actions, since the consequence of these actions cannot be presumed without additional information about the environment. One straightforward way to implement such a constraint is to explicitly model a given data distribution via behavior cloning and directly force a policy not to select uncertain actions.
  However, many offline RL methods instantiate the constraint indirectly---for example, pessimistic value estimation---due to a concern about errors when modeling a potentially complex behavior policy.
  In this work, we argue that it is not only viable but beneficial to explicitly model the behavior policy for offline RL because the constraint can be realized in a stable way with the trained model.
  We first suggest a theoretical framework that allows us to incorporate behavior-cloned models into value-based offline RL methods, enjoying the strength of both explicit behavior cloning and value learning. 
  Then, we propose a practical method utilizing a score-based generative model for behavior cloning. With the proposed method, we show state-of-the-art performance on several datasets within the D4RL and Robomimic benchmarks and achieve competitive performance across all datasets tested.
\end{abstract}

\section{Introduction}\label{intro}

The goal of offline reinforcement learning (RL) is to learn a policy purely from pre-generated data. 
This data-driven RL paradigm is promising since it opens up a possibility for RL to be widely applied to many realistic scenarios where large-scale data is available.

Two primary targets need to be considered in designing offline RL algorithms: maximizing reward and staying close to the provided data.
Finding a policy that maximizes the accumulated sum of rewards is the main objective in RL, and this can be achieved via learning an optimal Q-value function.
However, in the offline setup, it is often infeasible to infer a precise optimal Q-value function due to limited data coverage \cite{levine2020offRLtuto,MBS:Liu:neurips2020}; for example, the value of states not shown in the dataset cannot be estimated without additional assumptions about the environment. This implies that value learning can typically be performed accurately only for the subset of the state (or state-action) space covered by a dataset.
Because of this limitation, some form of imitation learning objectives that can force a policy to stay close to the given data warrants consideration in offline RL.

Recently, many offline RL algorithms have been proposed that instantiate an imitation learning objective without explicitly modeling the data distribution of the provided dataset. For instance, one approach applies the pessimism under uncertainty principle in value learning \cite{pessimism,CQL,fishercql:kostrikov:icml2021} in order to prevent out-of-distribution actions from being selected.
While these methods show promising practical results for certain domains, it has also been reported that such methods fall short compared to simple behavior cloning methods \cite{robomimic:Mandlekar:CoRL2021,implicitBC:florence:CoRL2021} which only model the data distribution without exploiting any reward information.
We hypothesize that this deficiency occurs because the imitation learning objective in these methods is only indirectly realized without explicitly modeling the data distribution (e.g. by pessimistic value prediction).
Such an indirect realization could be much more complicated than simple behavior cloning for some data distributions since it is often entangled with unstable training dynamics caused by bootstrapping and function approximation.
Hence, implicit methods are prone to over-regularization \cite{cql-workflow:kumar:corl2021} or failure, and they may require delicate hyperparameter tuning to prevent this deficiency \cite{rvs:emmons:ICLR2022}.
Yet, at the same time, it is obvious that simple behavior cloning cannot extract a good policy from a data distribution composed of suboptimal policies.

To this end, we ask the following question in this paper:
Can offline RL benefit from explicitly modeling the data distribution via behavior cloning no matter what kind of data distribution is given?
Previously, there have been attempts to use an explicitly trained behavior cloning model in offline RL~\cite{BRAC,BEAR,BCQ,MBS:Liu:neurips2020}, but we argue that two important elements are missing from existing algorithms. First, high-fidelity behavior cloning has not been achieved, despite the need in offline RL for precise estimation of behavior policy \cite{levine2020offRLtuto}.
Second, the trained behavior cloning models have only been utilized with heuristics or proxy formulations that are only empirically justified \cite{BRAC,BEAR}.
Therefore, we tackle these two problems by: first, incorporating a state-of-the-art score-based generative model \cite{score:song:neurips2019,scorev2:song:neurips2020,sde:song:iclr2021} to fulfill the high-fidelity required for offline RL,
and second, by proposing a theoretical framework, direct Q-penalization (DQP), that provides a mechanism to integrate the trained behavior model into value learning.
Furthermore, DQP can provide an integrated view of different offline RL algorithms, helping to analyze the possible failures of these algorithms. 

We evaluate our algorithm on various benchmark datasets that differ in quality and complexity, namely D4RL and Robomimic. Our method shows not only competitive performance across different types of datasets but also state-of-the-art results on complex contact-rich tasks, such as the transport and tool-hang tasks in Robomimic. The results demonstrate the practical effectiveness of the proposed algorithm as well as the advantage of explicit behavior cloning which was previously considered unnecessary or infeasible \cite{levine2020offRLtuto,Goo2021YOEO}.

To summarize, our contributions are: (1) We provide a theoretical framework for offline RL, DQP, which unifies previously disparate offline RL algorithms; (2) Using DQP, we suggest a principled offline RL formulation that incorporates an explicitly trained behavior cloning model; (3) We propose a practical algorithm that instantiates the above formulation, leveraging a score-based generative model; and (4) we achieve competitive and state-of-the-art performance across a variety of offline RL datasets.

\section{Related Works}\label{related-works}

Offline RL is about exploitation; the end goal of offline RL is to extract the best possible policy from a given dataset, regardless of the quality of the trajectories that compose the dataset \cite{levine2020offRLtuto}. One of the simplest approaches to tackle this problem is imitation learning (IL)~\cite{schaal1999imitation,implicitBC:florence:CoRL2021} hoping to recover the performance of the behavior policy used to generate the dataset. 
However, it is obvious that imitating behavior policy fails to achieve the end goal since one cannot outperform the behavior policy by just imitating it.
Furthermore, when a dataset contains trajectories from many policies, imitating the entire trajectories will not result in the performance of the best policy within the dataset.
To remedy this, recent approaches explore conditioned, filtered, or weighted imitation learning, which tries to imitate only good trajectories \cite{AWR,CRR:Wang:2020} or extracts actions that are related to a specific goal \cite{ding2019goal,ghosh2019learning} or rewards \cite{srivastava2019training,kumar2019reward}. The suggested methods are attractive especially for practitioners due to their conceptual and methodological simplicity, but it is often questioned whether such methods can work if a dataset is highly suboptimal \cite{rvs:emmons:ICLR2022} or requires trajectory stitching \cite{D4RL,cql-workflow:kumar:corl2021}. 

Another direction of research tackles offline RL problems via value learning, trying to resolve the distribution shift problem that arises in the offline setup. Since distribution shift commonly results in overestimation of values, the algorithms belonging to this category try to estimate values pessimistically for out-of-distribution inputs \cite{CQL,Goo2021YOEO}, sometimes by explicitly quantifying the certainty with a trained transition dynamics model \cite{MOPO,morel} or a generative model \cite{anti-exploration:rezaeifar2021offline,PLOFF:Dadashi:ICML21}. Intuitively, value-based offline RL methods are preferred over the behavior cloning-based counterpart since the value function could tell both what should and should not be done while imitation learning only utilizes the half side of the information. Also, several theoretical works \cite{whenshould:kumar:ICLR2022,MBS:Liu:neurips2020,bridging:rashidinejad:neurips2021} and empirical results \cite{D4RL,whenshould:kumar:ICLR2022} support the idea. At the same time, however, there is research arguing otherwise with empirical results \cite{robomimic:Mandlekar:CoRL2021,implicitBC:florence:CoRL2021,norlnosim:hahn:neurips2021,rvs:emmons:ICLR2022} that indicate the practical difficulties required in stabilizing the learning, such as delicate hyperparameter tuning \cite{cql-workflow:kumar:corl2021,rvs:emmons:ICLR2022}.

In this work, we present a new method that combines the strength of IL-based and value-based methods, which can be broadly categorized as a policy-constraint offline RL algorithm. In this family of algorithms, stable value learning is implemented by constraining a policy to be close to the behavior policy. Specifically, based on how the constraint is instantiated, it can be further categorized into implicit methods \cite{AWR,CRR:Wang:2020,IQL:kostrikov2021,onestep:brandfonbrener:neurips2021}, which constraint a policy via weighted behavior cloning, and explicit methods \cite{BCQ,BEAR,BRAC,MBS:Liu:neurips2020,TD3BC:fujimoto:neurips2021}, which constraint the policy learning via value penalty or policy regularization. We follow the overall learning structure of the explicit policy-constraint method, but we try to enhance its subpar performance by first proposing a theoretical framework that provides a principled way to regularize the policy and second using a powerful generative model for explicit behavior cloning.

\section{Preliminaries}

We use a Markov Decision Process (MDP) as a foundation for our mathematical framework. An MDP is defined by a tuple $\mathcal{M} = (\mathcal{S},\mathcal{A},T,d_0,r,\gamma)$; a set of states $s \in \mathcal{S}$, a set of actions $a \in \mathcal{A}$, transition dynamics $T = p(s'|s,a)$, an initial state distribution $d_0(s_0)$, a reward function $r(s,a)$, and a discount factor $\gamma$. In this setup, the goal of reinforcement learning is to find an optimal policy $\pi^*(a|s)$ that maximizes the expected sum of discounted reward (return) $J(\pi)$:
\begin{equation}
 \pi^* = \argmax_\pi J(\pi) = \argmax_\pi \mathbb{E}_{\tau \sim \rho_\pi} \left[ \sum_{t=0}^{H} \gamma^t r(s_t,a_t) \right],
\end{equation}
where $\tau$ is a sequence of states and actions $(s_0,a_0,\cdots,s_{H},a_{H})$ of length $H$, and $\rho_\pi$ is a trajectory distribution of a policy $\pi$, which can be represented as $\rho_\pi(\tau) = d_0(s_0)\prod_{t=0}^{H} \pi(a_t|s_t)T(s_{t+1}|s_{t},a_{t})$.

We can directly optimize the return when we can compute the gradients of $J(\pi_\phi)$ with respect to the policy parameters $\phi$ \cite{policygradient-williams1992,ppo:schulman2017proximal,trpo:schulman2015trust}, but this approach is not straightforwardly extend to an offline setting since on-policy data is typically required to compute the gradient. Instead, it is more common for offline RL methods to extend dynamic programming approaches which are formed around the action value function $Q^\pi(s,a)$ which is formally defined as: $Q^\pi(s_t,a_t) = \mathbb{E}_{\tau \sim p^\pi(s,a)} \left[ \sum_{t'=t}^{H} \gamma^{t'-t} r(s_{t'},a_{t'}) \right]$. The Q function of a certain policy $\pi_k$ always implies a greedy policy $\pi_{k+1}$, which is better than or equal to the evaluation target policy $\pi_k$. Therefore, an optimal policy can be found by iteratively evaluating a Q function for a new greedy policy $\pi_{k+1}$ until convergence:
\begin{equation}
  \begin{cases}
  Q^{\pi_k} = \lim_{n \rightarrow \infty} \big(\mathcal{B}^{\pi_k}\big)^n Q^{\pi_{k-1}} \quad \text{(policy evaluation)},\\
  \pi_{k+1}(a|s) = \argmax_a \ Q^{\pi_k}(s,a) \quad \text{(policy improvement)}, \label{policy-iteration}
  \end{cases}
\end{equation}
where $\mathcal{B}^\pi$ is the Bellman operator, which has the ground truth $Q^\pi$ as a unique fixed point \cite{lagoudakis2003least}. This algorithm is called policy iteration. Although the convergence of the algorithm is restricted to the scenario where the unique fixed point is reachable \cite{suttonbartoRL}, policy iteration has been widely used as a backbone for most offline RL algorithms due to its extensibility to off-policy data; policy evaluation can be done with off-policy data using bootstrapping. When the value function and the policy are represented with parameters $\theta$ and $\phi$ respectively, the policy iteration algorithm with bootstrapping has the following form:
\begin{equation}
  \begin{cases}
  \theta_{k+1} \leftarrow \argmin_{\theta} \mathbb{E}_{s,a,r,s' \sim D}
  \bigg[ d\big(Q_\theta(s,a), r + \gamma \mathbb{E}_{a' \sim \pi_k (a'|s')} Q_k(s',a') \big) \bigg]\\
  \phi_{k+1} \leftarrow \argmax_{\phi} \mathbb{E}_{s \sim D, a \sim \pi_\phi(a|s)} \bigg[ Q_{k+1}(s, a) \bigg]
  \end{cases} \label{policy-iteration-in-the-wild}
\end{equation}
where $k$ is an update step, $d$ is a distance metric such as squared $l_2$ or Huber loss, and $D$ is a provided (offline) dataset that contains transition tuples $D = {(s,a,r,s')}$. For the brevity of notation, we denote $Q_{\theta_{k}} := Q_k$ and $\pi_{\phi_{k}} := \pi_k$.

\section{Method}\label{method}

In this paper, we consider offline RL algorithms that utilize the policy iteration scheme shown in Eq.~\ref{policy-iteration-in-the-wild}, which covers several offline RL methods \cite{CQL,BEAR,BRAC}.
In this family of methods, the correctness of the value function becomes the major concern since the policy evaluation could diverge due to the data restriction imposed by the offline setup.
Specifically, divergence can happen because of the bootstrapping and the function approximation; when some estimates are erroneously high due to poor generalization, the over-estimated values are likely to be picked up on the policy improvement step and feed back to policy evaluation via bootstrapping, completing a vicious cycle that causes training to diverge.
Therefore, offline RL methods focus on solving the overestimation problem with different regularization methods, such as policy constraints \cite{BEAR,BRAC} or pessimism \cite{CQL,Goo2021YOEO}.

One straightforward solution to the over-estimation problem is directly penalizing Q estimation \cite{anti-exploration:rezaeifar2021offline,PLOFF:Dadashi:ICML21} with a penalty function $p(s,a)$: $\tilde Q_\theta(s,a) = Q_\theta(s,a) - p(s,a)$.
This can be a solution because this penalty function, if chosen carefully, can reduce erroneously high values and prevent them from being propagated via bootstrapping. We refer to this family of algorithms as Direct Q penalization (DQP).

In DQP, we can easily observe that the penalty function that describes the oracle Q estimation error (i.e. $p(s,a) = Q_{\theta} - Q^{\pi_\phi}$) is the best solution. Therefore, we want to design a penalty function that resembles the oracle error. Since the estimation error is likely to occur more often for out-of-distribution state-action pairs $(s,a') \notin D$, a few ad-hoc methods have been proposed for the purpose of measuring the epistemic uncertainty of $Q_{\theta}$; the aleatoric uncertainty of the transition dynamics model is suggested as a proxy for the epistemic uncertainty of $Q_\theta$ \cite{MOPO}, and generative models \cite{anti-exploration:rezaeifar2021offline} or pseudometrics \cite{PLOFF:Dadashi:ICML21} are proposed with the purpose of distinguishing whether a particular $(s,a)$ is in-distribution or not.

However, it has not been thoroughly investigated how penalty functions affect the policy iteration process, nor what penalty functions are best for offline RL. Without answers to these questions, DQP methods can only be understood as ad-hoc methods in which heuristically designed penalty functions are used to prevent the overestimation. To this end, we address the following questions: (1) What is the effect of direct Q-penalization in the context of the policy iteration framework?; (2) How can we design an appropriate penalty function based on this analysis?; (3) How can we instantiate the penalty function and achieve strong performance across different offline RL datasets?

\subsection{Theoretic Background on Direct Q-penalization}

We describe a theorem that answers the first question: soft policy iteration \cite{SAC} with a penalized value function $\tilde Q$ is equivalent to policy iteration regularized by $\KL\big(\pi(s)\|\pi_p(s)\big)$ where $\pi_p(a|s) = \softmax\big(-p(s,a)\big)$. This theorem is a generalized version of the theorem shown in \cite{anti-exploration:rezaeifar2021offline}, which does not require unnecessary assumptions on the penalty function.

\begin{nutheorem}[Equivalence between KL-policy regularization and DQP ]
  The following two algorithms are equivalent. \normalfont\\
  Policy iteration w/ KL-policy regularization:
  \begin{flalign*}
    &\begin{cases}
      \theta_{k+1} \leftarrow \argmin_{\theta} \mathbb{E}_{s,a,r,s' \sim D} \bigg[ d \Big( Q(s,a), r + \gamma \langle \pi_{k},Q_k \rangle(s') - \gamma \KL\big(\pi_{k}(s')\|\pi_p(s')\big)\Big)\bigg]\\
      \phi_{k+1} \leftarrow \argmax_{\phi} \mathbb{E}_{s\sim D} \bigg[ \langle \pi_\phi, Q_{k+1} \rangle(s) - \KL \big(\pi_\phi(s)\|\pi_p(s)\big) \bigg].
    \end{cases}&
  \end{flalign*}
  Soft policy iteration \cite{SAC} w/ penalty:
  \begin{flalign*}
    &\begin{cases}
      \theta_{k+1} \leftarrow \argmin_{\theta} \mathbb{E}_{s,a,r,s' \sim D} \bigg[ d \bigg( Q(s,a), r + \gamma \Big( \langle \pi_{k}, Q_k-p \rangle(s') - Z(s') + \entropy \big(\pi_{k}(s')\big) \Big) \bigg) \bigg]\\
      \phi_{k+1} \leftarrow \argmax_{\phi} \mathbb{E}_{s\sim D} \bigg[ \langle \pi_\phi, Q_{k+1}-p \rangle(s) + \entropy \big(\pi_\phi(s)\big) \bigg]
    \end{cases}&
  \end{flalign*}
  where $d$ is a distance metric, $\langle u_1,u_2 \rangle := \sum_a u_1(\cdot,a)u_2(\cdot,a)$, and $Z(s) = \ln \sum_{a} \exp\big(-p(s,a)\big)$.
  \label{theorem}
\end{nutheorem}
\begin{proof}
The common term in KL policy regularization can be rearranged as follows:
\begin{align*}
    \langle \pi, Q \rangle(s) - \KL\big(\pi(s)\|\pi_p(s)\big) &= \langle \pi, Q \rangle(s) - \langle \pi,\ln\pi - \ln\pi_p \rangle(s)\\
    &= \langle \pi, Q + \ln \pi_p\rangle(s) - \langle \pi,\ln \pi \rangle(s) \\ 
    &= \langle \pi, Q - p\rangle(s) - Z(s) + \entropy \big(\pi(s)\big).
\end{align*}
Then, we can get the equivalence when we replace the term with the rearranged term. Note that the normalization term $Z(s)$ can be dropped in the policy update step since $Z(s)$ is not a function of $\phi$. Also, we can safely ignore $Z(s)$ in the policy evaluation step when $|Z(s) - Z(s')| < \epsilon$ for any pair of $(s,s') \in \mathcal{S} \times \mathcal{S}$, because it does not affect the policy improvement step.
\end{proof}

While we show the equivalence for the case of applying the penalty function to both policy evaluation and improvement, a similar proof can be easily shown for the algorithm that applies the penalty only for the policy evaluation \cite{CQL, MOPO} or for policy improvement \cite{BEAR}. Therefore, we can compare different offline RL algorithms in terms of the penalty function that each algorithm uses. In Table~\ref{tab:others}, we summarize the penalty functions of some representative algorithms.

\begin{table}[t]
  \centering
  \caption{The penalty functions of different offline RL algorithms.}
  \label{tab:others}
  \begin{tabularx}{\textwidth}{llX}
  \toprule
  & $p(s,a)$ & Remark \\ \midrule
  BRAC-KL \cite{BRAC} & $- \log(\beta(a|s))$ &        \\
  BRAC-MMD$^2$ \cite{BEAR, BRAC} & $\text{MMD}^2(\pi_k, \beta)$ &\\
  Anti-Exploration \cite{anti-exploration:rezaeifar2021offline} & $\alpha |a - \text{Dec}(\text{Enc}(s,a))|_2^2$ & $\text{Enc}$ and $\text{Dec}$ are CVAE \cite{vae:kingma2013}.\\
  PLOFF \cite{PLOFF:Dadashi:ICML21}  & $\alpha_1 Q_k(s,a) \exp (-\alpha_2 \ \text{D}(s,a))$ & $\text{D}$ is pseudometric \cite{PLOFF:Dadashi:ICML21}. \\
  MOPO \cite{MOPO}    & $\alpha |\Sigma(s,a)|$ & $\Sigma$ is the std. of the trained $\mathcal{T}$.\\
  CQL \cite{CQL} & $\alpha_k [ \frac{\mu_k}{\beta} -1 ]$ & $\mu_k$ is the soft-policy given $Q_k$. \\
  \bottomrule
  \end{tabularx}
\end{table}

\subsection{What makes a good penalty function?}

Theorem~\ref{theorem} shows the connection between a penalty function and its effect as a policy regularizer, and can help to guide the construction of an effective, principled penalty function. 
Specifically, we propose a penalty function that can instantiate the support set constraint \cite{BEAR,MBS:Liu:neurips2020}, which restricts the action space of a trained policy to be in the support set of the behavior policy $\beta(a|s)$.
The support set constraint is an effective way to solve the offline RL problem in that the suboptimality caused by the constraint is bounded \cite{BEAR,MBS:Liu:neurips2020}. While the previous works express the constraint in terms of the distribution constrained Bellman operator, we represent the constraint via the penalty function since it allows us to compare different offline RL algorithms under the same viewpoint.

The following penalty function instantiates the support set constraint:
\begin{equation}
p(s,a) = \begin{cases}
  0\ \text{for}\ \{(s,a)|\beta(a|s) \ge \epsilon \}  \\
  \infty  \ \text{otherwise}
  \end{cases}
\end{equation}
where $\epsilon$ is a threshold hyperparameter to decide whether $(s,a)$ is considered out-of-support or not.
The penalty function carries the same effect as the filtration operator in \cite{MBS:Liu:neurips2020} under the policy iteration scheme. 
This is because the function prevents out-of-support actions from being chosen by the policy while it imposes no preference over in-support actions; a rare action that has not occurred often in a dataset can be selected as long as it provides a high Q value. The indifference is desirable property when good trajectories compose only a small portion of a dataset since good actions could be swamped by more frequent actions if the penalty function is designed to prefer more frequent actions.

We can also confirm the characteristics of the penalty function by observing the flip side: the penalty-induced policy $\pi_p$ and the KL-constraint $\KL(\pi \| \pi_p)$. Since the reverse KL term makes $\pi$ seek a mode of $\pi_p$ which is the uniform distribution for the in-support actions, the policy $\pi$ is guided to select one of the actions in the support set while there is no preference over actions in the set. Therefore, the penalty function instantiates the support set constraint.

Given the proposed penalty function, the similarity between different offline RL methods can be observed. For instance, we can see that BRAC-KL and CQL \cite{CQL} penalizes the out-of-support actions infinitely: when $\beta(a|s)$ is zero, the penalty becomes infinite.
However, some discrepancies can also be noted, and this provides some hints on how and why other methods could fail to improve performance over given datasets or to prevent out-of-support actions.
First, BRAC-KL could fail because it prefers actions that are more frequently executed by the behavior policy.
It would make a policy to seek the mode of $\beta(a|s)$ when the KL regularization term dominates the policy update step.
Therefore, the algorithm could work like behavior cloning, and the algorithm could fail to fully exploit the provided data.
CQL could also exhibit a similar problem since the penalty function is defined with $\beta$; when CQL is tuned to strongly penalize the out-of-support action (i.e., $\alpha_k$ is large), it could constrain a policy too harshly, letting the policy nothing but mimic the dataset \cite{cql-workflow:kumar:corl2021}.
The formulation of BEAR \cite{BEAR}, on the other hand, can avoid the problem of BRAC-KL and CQL by indirectly utilizing the behavior policy $\beta$ in defining the penalty function; they use MMD distance between $\pi$ and $\beta$ to instantiate the support set constraint. However, their use of MMD distance for the constraint is only empirically justified with certain conditions, such as the small number of samples in computing the distance \cite{BEAR}.
Another common problem that arises in other penalty functions is their use of proxy and their formulation; for example, a conditional variational autoencoder (CVAE)\cite{anti-exploration:rezaeifar2021offline}, a pseudometric \cite{PLOFF:Dadashi:ICML21}, or a transition dynamics model \cite{MOPO} are estimated instead of the behavior policy $\beta$, and penalty functions are designed heuristically with the proxy estimates. While such formulations could show some positive correlation to the suggested penalty function, there is no clear connection that allows us to interpret the penalty in terms of $\beta$ or the support set.

\subsection{Practical Implementation}

We now propose a practical algorithm that instantiates the penalty function designed above. Essentially, the designed penalty function serves to determine whether an action $a$ at a certain state $s$ is likely to be executed by the behavior policy $\beta$. Therefore, we can implement the penalty function simply by cloning the behavior policy explicitly and checking the likelihood $\beta(a|s)$ with the cloned model.

\begin{figure}
  \centering
  \includegraphics[width=1.0\textwidth]{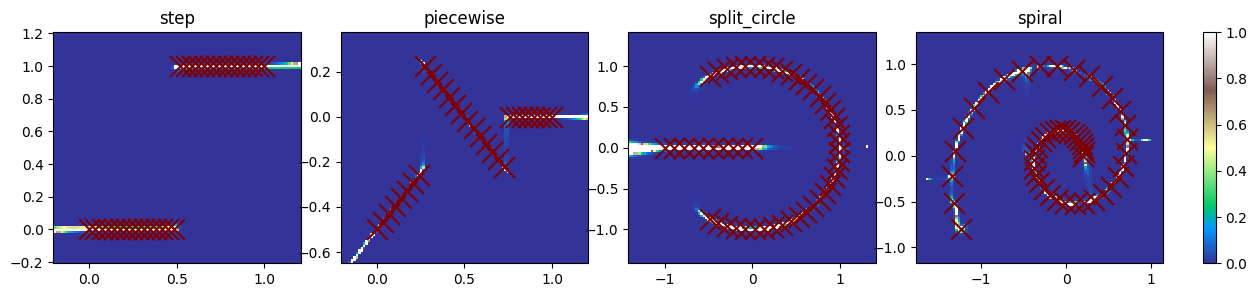}
  \caption{The estimated probability for complex, discontinuous, and multi-modal conditional distributions $\beta(a|s)$. The state and action spaces are one-dimensional, and the conditional probability estimated using score-based generative models $s_\psi$ is represented with a color map. Red x marks represent the training samples.
  }
  \label{sgm-test}
  \vspace{-1em}
\end{figure}

There have been other research works that have tried to model a behavior policy using generative models, such as variational auto-encoders (VAEs)~\cite{BCQ,anti-exploration:rezaeifar2021offline}. However, the performance of these approaches is limited compared to methods that do not explicitly clone the behavior policy $\beta$.
We presume that the reason for this failure is the limited expressivity of the generative model; since the behavior policy $\beta$ can be complex, discontinuous, and multi-modal, only a very expressive model can successfully model the policy. To this end, we chose to use a score-based generative model \cite{score:song:neurips2019,scorev2:song:neurips2020,sde:song:iclr2021}, which has recently shown great success in generating high-quality images.
Furthermore, the score-based generative model allows an exact likelihood computation which is essential in instantiating the penalty function. 
We briefly examine the ability of the score-based generative model using four discontinuous multi-modal distributions, and the results are shown in Figure~\ref{sgm-test}. In all four cases, the inferred probability distribution is very sharp, and its log probability resembles the penalty function we proposed.

In the score-based generative model, a target distribution $p(x)$ is indirectly expressed and trained in the form of the gradient of a log probability density function $\nabla_x \log p(x)$, often referred to as the (Stein) score function \cite{liu2016kernelized}. When we train an accurate score function $s_\psi(s,a)$ of the conditional distribution $\beta(a|s)$ via score-matching algorithm \cite{score:song:neurips2019}, we can instantiate the penalty function with a hyperparameter $\epsilon$ which decides whether a certain action $a$ given $s$ is considered to be sufficiently in-support or not. While this formulation allows direct instantiation of the penalty function that can be plugged into the DQP framework, it is computationally prohibitive. This is because, first, we need to run an iterative algorithm to compute the log-likelihood from the score function, and second, it could hurt the generalization performance of the value function since Q has to output an extremely wide range of values including negative infinity. Therefore, we propose a practical approximation of the policy iteration algorithm that utilizes the proposed penalty function.

The key observation is that the policy trained on top of the penalized value function will never select out-of-support actions due to penalization. This allows two modifications to the original policy iteration algorithm: First, we can perform policy evaluation only considering in-support state-action pairs; i.e., we can bootstrap from one of the samples from $\beta(a|s)$, and we can skip the penalty computation since the penalty is zero for in-support data-points under the suggested penalty function. While sampling using the score function also requires expensive iterative computation, we can greatly reduce the computation by prepopulating samples for states that exist in the dataset and repeatedly using it in the policy evaluation step. Specifically, the policy evaluation is done with the following loss function:
\begin{equation}
  L_{\text{policy-eval}} = \mathbb{E}_{s,a,r,s' \sim D} \bigg[ d\bigg( Q_\theta(s,a), r + \gamma \Kth_{a' \in \supp (\beta(s'))} \big[ Q_{\tilde{\theta}} (s',a') \big] \bigg) \bigg]
  \label{ARQ-eval}
\end{equation}
where $Q_{\tilde{\theta}}$ is a slowly updated target network, and $\Kth$ is an operator that selects the $K$-th item among candidates. When $K=1$, it becomes $\max$ operator. Both $Q_{\tilde{\theta}}$ and $\Kth$ operators are adapted to stabilize the learning.

Second, we can skip the policy improvement step since policy evaluation is done with pre-generated samples, not depending on any parameterized policy. Instead, we can define an implicit policy using the last $Q_\theta$ and $s_\psi$:
\begin{equation}
\pi(a|s) = \frac{\exp\big(\alpha Q_\theta(s,a)\big)}{\sum_{a' \in \text{supp}(\beta)} \exp\big(\alpha Q_\theta(s,a')\big)} \quad\text{or}\quad 
\frac{\exp\big(\alpha A(s,a)\big)}{\sum_{a' \in \text{supp}(\beta)} \exp\big(\alpha A(s,a')\big)}
\end{equation}
where $A(s,a) = Q_\theta(s,a) - \frac{1}{|\text{supp}(\beta)|}\sum_{a' \in \text{supp}(\beta)} Q_\theta(s,a')$ is an advantage function, and $\alpha$ is a temperature parameter that controls the policy softness with regard to the $Q$ or $A$; when $\alpha$ is zero or infinity, the resulting policy becomes $\beta(a|s)$ or greedy with regard to $Q_\theta$, respectively. To sample an action from the policy $\pi$, we can sample one action from an empirical action distribution consisting of samples generated from $\beta$ on the fly. Alternatively, we can also train parameterized policy $\pi_\phi(a|s)$ using advantage-weighted regression (AWR) \cite{AWR} with the advantage function $A(s,a)$.

The resulting algorithm can be regarded as one special type of Q-learning in which we restrict the domain of the maximum operator to in-support actions. We refer to this algorithm as Action-Restricted Q-learning (ARQ). The pseudocode of ARQ is shown in Algorithm~\ref{alg}. Also, ARQ can be seen as an extension of MBS-QI \cite{MBS:Liu:neurips2020} in that it makes the existing algorithm applicable to MDP with a continuous action space. Note that ARQ is one way to instantiate the penalty function, mainly due to the expensive computational cost of the score-based generative model. We discuss about other possible instantiations in the discussion section.
\begin{figure}
\centering
\begin{minipage}{0.85\linewidth}
\begin{algorithm}[H]
  \SetKwInOut{Input}{Input}
  \Input{Dataset $D = \{(s,a,r,s')\}$, Hyperparameter $N$, $\epsilon$, $K$, $\alpha$}
  Initialize $s_\psi(a|s)$, $Q_\theta(s,a)$, and $\pi_\phi(a|s)$ (if needed) \\
  Train $s_\psi$ with a score matching algorithm \cite{sde:song:iclr2021} \\
  Sample $N$ in-support actions for $s \in D$ (i.e., $\beta_\psi(a|s) > \epsilon$)\\
  \While{until convergence}{
      Update $\theta$ with $\nabla_\theta L_{\text{policy-eval}}$ (Eq.~\ref{ARQ-eval})
  }
  \While{until convergence}{
      Update $\phi$ with $\nabla_\phi - \mathbb{E}_{s,a \sim D} \big[e^{\alpha A(s,a)} \log \pi_\phi(a|s) \big]$; AWR~\cite{AWR}
  }
  \Return $s_\psi$, $Q_\theta$, and $\pi_\phi$
  \caption{Action-Restricted Q-learning (ARQ)}
  \label{alg}
\end{algorithm}
\end{minipage}
\vspace{-1em}
\end{figure}

\section{Experiments}\label{expr}

Our empirical goal is to design an algorithm that enjoys the strength of both explicit behavior cloning and value learning. Therefore, the main goal of the experiments is to check whether the proposed algorithm ARQ achieves competitive performance on different types of datasets, ranging from a dataset that consists of near-optimal data in which explicitly cloning a behavior is adequate, to a dataset containing various suboptimal trajectories in which learning a value function is necessary.

The implementation of ARQ consists of four steps: score-based generative model $s_\psi$ learning, sampling, Q-learning $Q_\theta$, and optional explicit policy $\pi_\phi$ training.
As for the hyperparameters, we tune the hyperparameters $K$ and $\alpha$ for each group of datasets using random search while we use $N=30$ and $\epsilon=e^{-5}$ (i.e., 30 samples are generated and dropped if the likelihood is lower than $\epsilon=e^{-5}$) all across the datasets tested.
For the detailed implementation details and hyperparameters, please refer Appendix~\ref{app:impl} or the provided code
\footnote{\url{https://github.com/hiwonjoon/ARQ-public}}.

The proposed method is evaluated on various simulated benchmark datasets from simple low-dimensional locomotion tasks to complex contact-rich manipulation tasks. Specifically, we use locomotion~\cite{openaigym}, Adroit~\cite{adroit}, Kitchen~\cite{frankakitchen}, and Antmaze tasks in D4RL~\cite{D4RL} (licensed under CC BY), and six manipulation tasks in Robomimic~\cite{robomimic:Mandlekar:CoRL2021} (disclosed with MIT License). We use medium-replay, medium, expert, and medium-expert datasets of the locomotion task.
We use machine-generated (mg.), proficient-human (ph.), and multi-human (mh.) datasets of Robomimic, each of which consists of a replay buffer of an SAC training run, trajectories of a proficient human demonstrator, and trajectories of multiple human demonstrators with different levels of proficiency.

We compare the proposed method to behavior cloning baselines, specifically ordinary BC and implicit behavior cloning~\cite{implicitBC:florence:CoRL2021}, and prior state-of-the-art offline RL methods. Namely, we compare the performance of our method with TD3+BC~\cite{TD3BC:fujimoto:neurips2021}, Decision Transformer (DT) \cite{DecisionTransformer:chen:neurips2021}, One-step RL~\cite{onestep:brandfonbrener:neurips2021}, CQL~\cite{CQL}, and IQL~\cite{IQL:kostrikov2021}.
The aggregated results are displayed in Table~\ref{tab:short}.

\begin{table}[t]
  \centering
  \caption{Aggregated performance of prior methods and ours on D4RL\cite{D4RL} and Robomimic\cite{robomimic:Mandlekar:CoRL2021} datasets. Each number represents the performance relative to a random policy as $0$ and an expert policy as $100$. Unless noted as (ours) or (repro.), all the numbers are borrowed from \cite{IQL:kostrikov2021,TD3BC:fujimoto:neurips2021,robomimic:Mandlekar:CoRL2021}. The numbers generated by us are averaged over 3 different random seeds. We run IQL on Robomimic by ourselves using the author-provided implementation. The full results are available in the appendix.
  }
  \label{tab:short}
  \resizebox{\textwidth}{!}{%
  \begin{tabular}{@{}cccccccccccc@{}}
  \toprule
   & \multicolumn{3}{c}{Without reward} & \multicolumn{2}{c}{With $Q^\beta$} & \multicolumn{6}{c}{With reward / value function} \\
   & BC 
   & \begin{tabular}[c]{@{}c@{}}Impl.\\ BC\end{tabular} 
   & \begin{tabular}[c]{@{}c@{}}$s_\psi$\\ (ours)\end{tabular} 
   & \begin{tabular}[c]{@{}c@{}}One-\\ step\end{tabular} 
   & \begin{tabular}[c]{@{}c@{}}Q($\beta$)\\ +$s_\psi$\\ (ours)\end{tabular} 
   & DT 
   & \begin{tabular}[c]{@{}c@{}}TD3\\ +BC\end{tabular}
   & CQL
   & IQL 
   & \begin{tabular}[c]{@{}c@{}}ARQ\\ +$\pi_\phi$\\ (ours)\end{tabular} 
   & \begin{tabular}[c]{@{}c@{}}ARQ\\+$s_\psi$\\ (ours)\end{tabular} \\ \toprule
  \multicolumn{1}{l}{\textbf{locomtion-v2 (total)}} & 739 & 521 & 639 &  & 911 &  & \textbf{992} & \textbf{996} & \textbf{1007} & \textbf{1000} & 947 \\ \midrule
  \multicolumn{1}{l}{\textbf{adroit-v0 (total)}} & 105 &  & 116 &  & 160 &  &  & 94 & 118 & 95 & \textbf{161} \\ \midrule
  human-v0 (total) & 67 & \textbf{99} & 87 & - & 89 & - & - & 52 & 77 & 45 & 90 \\
  cloned-v0 (total) & 37 & - & 29 & 62 & \textbf{71} & - & - & 42 & 41 & 50 & \textbf{71} \\ \midrule
  \multicolumn{1}{l}{\textbf{kitchen-v0 (total)}} & 155 & 160 & 170 &  & 186 &  &  & 145 & 160 & 126 & \textbf{204} \\ \midrule
  complete & 65 & \textbf{85} & 74 & - & 75 & - & - & 44 & 63 & 37 & 77 \\
  partial & 38 & 38 & 45 & - & 59 & - & - & 50 & 46 & 50 & \textbf{70} \\
  mixed & 52 & 38 & 51 & - & 52 & - & - & 51 & 51 & 39 & \textbf{57} \\ \midrule
  \multicolumn{1}{l}{\textbf{antmaze-v0}} & 100 &  & 121 & 125 & 215 & 112 & 164 & 304 & 378 & \textbf{416} & 327 \\ \midrule
  umaze & 55 & - & 58 & 64 & 81 & 59 & 79 & 74 & 88 & \textbf{97} & 94 \\
  umaze-div. & 46 & - & 61 & 61 & 62 & 53 & 71 & \textbf{84} & 62 & 62 & 58 \\
  med.-play & 0 & - & 1 & 0 & 25 & 0 & 11 & 61 & 71 & \textbf{80} & 69 \\
  med.-div. & 0 & - & 1 & 0 & 45 & 0 & 3 & 54 & 70 & \textbf{82} & 65 \\
  large-play & 0 & - & 0 & 0 & 1 & 0 & 0 & 16 & \textbf{40} & \textbf{37} & 18 \\
  large-div. & 0 & - & 0 & 0 & 1 & 0 & 0 & 15 & 48 & \textbf{58} & 23 \\ \midrule
  \multicolumn{1}{l}{\textbf{D4RL (total)}} & 1,099 &  & 1,046 &  & 1,472 &  &  & 1,538 & \textbf{1,663} & \textbf{1,637} & \textbf{1,639} \\ \toprule
   & BC$^\star$ &  & \begin{tabular}[c]{@{}c@{}}$s_\psi$\\ (ours)\end{tabular} &  & \begin{tabular}[c]{@{}c@{}}Q($\beta$)\\ +$s_\psi$\\ (ours)\end{tabular} &  & BCQ$^\star$ & CQL$^\star$ & \begin{tabular}[c]{@{}c@{}}IQL\\ (repro.)\end{tabular} & \begin{tabular}[c]{@{}c@{}}ARQ\\ +$\pi_\phi$\\ (ours)\end{tabular} & \begin{tabular}[c]{@{}c@{}}ARQ\\ +$s_\psi$\\ (ours)\end{tabular} \\ \toprule
  \multicolumn{1}{l}{\textbf{robomimic (total)}} & 701 &  & 644 &  & \textbf{749} &  & 592 & 281 & 342 & 659 & \textbf{750} \\ \midrule
  mg.-lift & 65 &  & 29 &  & 86 &  & \textbf{91} & 64 & 79 & 79 & 82 \\
  mg.-can & 65 &  & 19 &  & 55 &  & \textbf{75} & 1 & 62 & \textbf{76} & 60 \\
  ph.-lift & \textbf{100} &  & \textbf{99} &  & \textbf{100} &  & \textbf{100} & 93 & 58 & \textbf{100} & \textbf{98} \\
  ph.-can & \textbf{95} &  & \textbf{95} &  & \textbf{93} &  & 89 & 38 & 26 & \textbf{92} & \textbf{95} \\
  ph.-square & \textbf{79} &  & 66 &  & 72 &  & 50 & 5 & 24 & 44 & 69 \\
  ph.-transport & 17 &  & \textbf{27} &  & \textbf{28} &  & 7 & 0 & 1 & \textbf{29} & \textbf{30} \\
  ph.-toolhang & 29 &  & \textbf{70} &  & 64 &  & 0 & 0 & 3 & 3 & \textbf{71} \\
  mh.-lift & \textbf{100} &  & \textbf{96} &  & \textbf{94} &  & \textbf{100} & 57 & 51 & \textbf{99} & \textbf{95} \\
  mh.-can & \textbf{86} &  & \textbf{84} &  & \textbf{89} &  & 63 & 22 & 25 & \textbf{90} & \textbf{86} \\
  mh.-square & \textbf{53} &  & 44 &  & \textbf{51} &  & 14 & 1 & 12 & 31 & \textbf{51} \\
  mh.-transport & 11 &  & \textbf{15} &  & \textbf{17} &  & 3 & 0 & 0 & \textbf{16} & 13 \\ \midrule
  \multicolumn{1}{l}{\textbf{D4RL + robomimic}} & 1,799 &  & 1,690 &  & 2,221 &  & 592 & 1,818 & 2,005 & 2,296 & \textbf{2,389} \\ \bottomrule
  \multicolumn{12}{r}{$\star$ represents that the best performance during training iterations is picked a posteriori.}
  \end{tabular}%
  }
  \end{table}

The proposed algorithm ARQ shows competitive performance on all ranges of datasets, from near-optimal ones in which simple behavior cloning is sufficient, to suboptimal datasets in which value learning is necessary.
Also, ARQ exhibits state-of-the-art performance on complex and contact-rich tasks, such as adroit, kitchen, and Robomimic datasets. The results indicate the practical effectiveness of the proposed algorithm, as well as the advantage of performing behavior cloning explicitly with high-fidelity models. 

To examine the importance of each component in ARQ, we run two ablations; we evaluate an implicit policy defined only with $s_\psi$ without any value function, and an implicit policy incorporating $Q^\beta$ instead of ARQ.
The results of the ablations affirm our hypotheses. First, when the dataset is near-optimal (e.g., adroit-human or proficient-human datasets), explicitly modeling the behavior policy can address the problem, and similar performance is obtained when we use ARQ. Next, we confirm the necessity of value learning and the ability of ARQ in leveraging the explicitly cloned behavior model in learning a value function. Especially in the tasks where trajectory stitching is required (e.g., kitchen and antmaze datasets), we can see performance improvement from $s_\psi$ and $Q(\beta)+s_\psi$ to ARQ$+s_\psi$, and we achieve  state-of-the-art performance with the help of the explicit models.

It is also noteworthy that the ablated method with $Q^\beta$ shows competitive results on a large number of benchmarks. Echoing prior research \cite{Goo2021YOEO,onestep:brandfonbrener:neurips2021}, these results indicate that the vast majority of offline RL benchmarks can be resolved without iterative value learning, while most offline RL algorithms tackle problems that arise from it. Therefore, in order to fairly evaluate the offline RL algorithms and thereby foster the advance of the offline RL field, it is essential to focus on environments that require value learning (e.g., antmaze) or develop new benchmarks.

\section{Discussion}\label{discussion}

We investigate an offline RL algorithm that combines explicit behavior cloning and value learning. We provide a theoretical framework, DQP, which enables various offline RL algorithms to be expressed in terms of different penalty functions, and we derive a principled penalty function that can leverage a behavior cloning model. Then, we provide a practical algorithm, ARQ, which realizes the derived penalty function. We implement the algorithm with a powerful generative model to maximize the full potential of ARQ. As a result, the proposed algorithm shows competitive results on most of the D4RL and Robomimic benchmarks and yields state-of-the-art results in several tasks. This indicates that the common presumption---that it is unnecessary or infeasible to estimate a behavior policy in offline RL---is likely incorrect.

The major drawback of the proposed algorithm is the computationally expensive sampling procedure. While it needs to be computed only once before the value learning step, it can take several hours to generate samples (90K samples are generated per hour on our in-house workstation with an Nvidia GTX 1080 Ti). Therefore, future research may examine how to reduce computational burden, for instance, by using different generative models for behavior cloning. Or, it may be possible to devise a method that directly utilizes the score function under the actor-critic framework; since the actor update step with the penalized Q function ($\tilde{Q} = Q - p$) only requires a gradient of $p$, not an exact penalty value,  the computational bottleneck may be avoided if the gradient of the penalty function can be computed directly from the score function. 

By using offline RL approaches such as the proposed method, RL can be applied to real-world tasks for which it was not previously applicable due to exploration risks, such as autonomous driving or healthcare. Although our work makes the learning process safer by training a policy fully offline, it could be still harmful to deploy a trained policy in the real world without thoroughly analyzing the possible errors the policy can make.
Therefore, it is also critical to determine whether a learned policy is well-aligned with human values \cite{brown2021value}.

\begin{ack}
This work has taken place in the Personal Autonomous Robotics Lab (PeARL) at The University of Texas at Austin. PeARL research is supported in part by the NSF (IIS-1724157, IIS-1638107, IIS-1749204, IIS-1925082), ONR (N00014-18-2243), AFOSR (FA9550-20-1-0077), and ARO (78372-CS).
This research was also sponsored by the Army Research Office under Cooperative Agreement Number W911NF-19-2-0333. The views and conclusions contained in this document are those of the authors and should not be interpreted as representing the official policies, either expressed or implied, of the Army Research Office or the U.S. Government. The U.S. Government is authorized to reproduce and distribute reprints for Government purposes notwithstanding any copyright notation herein.
\end{ack}

%% file: _appendix.tex
\section{Appendix}

\subsection{Score-based Generative Model \texorpdfstring{\cite{score:song:neurips2019,scorev2:song:neurips2020,sde:song:iclr2021}}{}}

In the score-based generative model, a target distribution $p(x)$ is indirectly expressed and trained in the form of the gradient of a log probability density function $\nabla_x \log p(x)$, often referred to as the (Stein) score function \cite{liu2016kernelized}.
This method circumvents several problems of other generative models. The main advantages are, first, we can circumvent the problem of the inferring normalizing constant that arises in likelihood-based methods \cite{vae:kingma2013,lecun2006tutorial}, and second, we can train the score function without worrying about training instability that arises in adversarial training \cite{goodfellow2014generative} via score-matching algorithms \cite{hyvarinen2005estimation, vincent2011connection}; the score-matching algorithm minimizes the gap between ground-truth score function and the estimates: $L_{\text{score-matching}} = \mathbb{E}_{p(x)}[\| \nabla_x \log p(x) - s_\psi(x)\|]^2_2]$. Since the objective is essentially regression with $l_2$ loss, the loss function does not require any assumptions on the parameterized function $s_\psi$, unlike the energy-based model \cite{lecun2006tutorial} in which strong regularization is often required for stable training.
These advantages make it possible to model a complex behavior policy $\beta$ with the high fidelity that offline RL requires.
We briefly examine the ability of the score-based generative model using four discontinuous multi-modal distributions, and the results are shown in Figure~\ref{sgm-test}. In all four cases, the inferred probability distribution is very sharp, and its log probability resembles the penalty function we proposed.

\subsection{Implementation Details}\label{app:impl}

The implementation of ARQ consists of four steps: score-based generative model $s_\psi$ learning, sampling, Q-learning, and optional explicit policy $\pi_\phi$ training. 
For the score-based generative model, sampling, and likelihood computation, we generally follow the implementation of \cite{sde:song:iclr2021} which trains a time-dependent neural network that approximates the reverse-time stochastic differential equations (SDE); a neural network learns to reverse the progressive diffusion process that turns a data point into random noise.
Specifically, we use a value-preserving SDE (VPSDE) with a neural network having residual connections \cite{he2015deep} with the embedding size of 256, and we stack 3 residual blocks. We use \texttt{swish} for nonlinearity.
Since our target distribution is conditional (i.e., conditioned on a state) unlike the original formulation, we extend a time-dependent neural network to be a function of a state, an action, and time.
We train the network with Adam with a learning rate of 1e-4 and batch size of 512, and we apply an exponential moving average with an average coefficient of 0.999.
Since the number of training data samples varies across datasets, we tested the different number of training iterations and ensembles to prevent overfitting. For training iterations, we tested 150,000, 300,000, and 1 million steps, and for ensemble, we tried single and 3 ensemble models. For ensemble training, we train each model with different random samples from the same data pool. The dataset-specific hyperparameters are shown in Table~\ref{tab:hpA}.

\begin{table}[h]
    \centering
    \caption{Dataset-specific hyperparameters used in training a score function $s_\psi$.}
    \label{tab:hpA}
    \begin{tabular}{@{}lcccccc@{}}
    \toprule
                        & \multicolumn{4}{c}{D4RL}                   & \multicolumn{2}{c}{Robomimic} \\
                        & Locomotion & Kitchen & Adroit  & AntMaze   & MG             & PH, MH       \\ \midrule
    Training Iterations & 1,000,000  & 300,000 & 300,000 & 1,000,000 & 1,000,000      & 150,000      \\
    \# Ensembles         & 1          & 1       & 3       & 1         & 3              & 3            \\ \bottomrule
    \end{tabular}
\end{table}

The second step of the ARQ algorithm is prepopulating samples for value learning. 
We use a predictor-corrector algorithm \cite{sde:song:iclr2021} to generate samples by solving a reverse SDE. Again, we followed the implementation of \cite{sde:song:iclr2021}, which uses the Euler-Maruyama method as a predictor and Langevin dynamics as a corrector. We discretize the time domain [1e-3,1] of $s_\psi$ into 500 steps, and we execute a single corrector step for every predictor step. For Langevin dynamics, we dynamically adjust the noise scale using the norm of the score; we use a score-to-noise ratio of 0.16 as used in \cite{sde:song:iclr2021}.
Since the suggested penalty function is defined based on its probability $\beta(a|s)$, we need to compute the likelihood of generated samples using $s_\psi$. For the likelihood computation, we use an instantaneous change-of-variable formula on top of the ordinary differential equation induced from the SDE. To solve the inverse problem of ODE, we use the RK45 algorithm of scipy.
For the detailed formulation and algorithmic detail regarding SDE, we refer to the original paper \cite{score:song:neurips2019,scorev2:song:neurips2020,sde:song:iclr2021} or our code.
We use $\epsilon=e^{-5}$ across all experiments, and we drop any samples that show a lower probability than the given threshold $\epsilon$. We generate 30 samples for both ARQ and the implicit policy using $s_\psi$.

For ARQ training, we use an MLP with 2 layers and 256 activation nodes to parameterize the value function $Q_\theta$, and we apply \texttt{ReLU} nonlinear activation. Also, we shape the reward function of datasets following \cite{IQL:kostrikov2021}; for the locomotion tasks, the reward is normalized by multiplying the ratio of returns between the worst and the best trajectories in the dataset, and for the antmaze tasks, the reward is set to -1 except the goal state. Similarly, we densify the reward function of Robomimic using the same technique used for the antmaze.
For stability, we train two Q functions and a slowly moving target network with a polyak coefficient of 0.995. We train 1 million timesteps using Adam optimizer and batch size of 512.
We perform a rough random search with the following range of values for the following hyperparameters: $K \in [3,6,9]$, a learning rate $\in$ [3e-4,1e-4]. The chosen hyperparameters are shown in Table~\ref{tab:hp-arq}.
For the $Q^\beta$ used in the ablation study, we use the same parameterization of the value function, but we train a single Q function with a slowly moving target network. The value function is trained 1 million time steps using Adam with a learning rate of 1e-4 and batch size of 512.

\begin{table}[h]
    \centering
    \caption{Dataset-specific hyperparameters used or in ARQ.}
    \label{tab:hp-arq}
    \begin{tabular}{@{}lcccccc@{}}
    \toprule
                  & \multicolumn{4}{c}{D4RL}                & \multicolumn{2}{c}{Robomimic} \\
                  & Locomotion & Kitchen & Adroit & AntMaze & MG           & PH, MH         \\ \midrule
    Learning rate & 3e-4       & 1e-4    & 1e-4   & 3e-4    & 1e-4         & 1e-4           \\
    $\Kth$        & 9          & 9       & 9      & 3       & 9            & 9              \\
    Reward & Normalized \cite{IQL:kostrikov2021} & Original & Original & -1 except goal \cite{IQL:kostrikov2021} & \multicolumn{2}{c}{-1 except goal} \\ \bottomrule
    \end{tabular}
\end{table}

For the implicit policy that is based on the samples of $s_\psi$, we first generate 30 samples using $s_\psi$, then we resample an action from the categorical distribution that treats advantages as logits.
Similarly, for the explicit policy, we train a policy using weighted behavior cloning \cite{AWR} where the weight is computed using the advantage.
We use state-independent stochastic policy used in \cite{IQL:kostrikov2021} for the locomotion, kitchen, and adroit tasks of D4RL datasets, which predicts the mean $\mu(a|s)$ and the state-independent standard deviation $\sigma(a)$ of a Gaussian distribution. We use a 2-layer MLP having 256 hidden units with \texttt{ReLU} activations.
For the antmaze tasks, we use the deterministic policy that omits the standard deviation prediction.
Similarly, a deterministic policy is used for the Robomimic datasets, but we use dense Resnet blocks for the parameterization. We stack 4 ResNet blocks, each of which has 2048 embedding dimensions. We tried Gaussian Mixture Network (GMM) as suggested in \cite{robomimic:Mandlekar:CoRL2021}, but we could not replicate the reported performance in the BC setting.
For D4RL and Robomimic, we train a policy for 1 million and 300,000 steps respectively, using the Adam optimizer with a learning rate of 3e-4.
The key hyperparameter for the policy is the temperature term $\alpha$. We tested the following range of values: [0.1, 1.0, 10.0, 30.0], and we display the chosen values in Table~\ref{tab:policy} along with other hyperparameters.

\begin{table}[h]
    \centering
    \caption{Dataset-specific hyperparameters used in the implicit policy and the explicit policy $\pi_\phi$.}
    \label{tab:policy}
    \begin{tabular}{@{}lcccccc@{}}
    \toprule
                                      & \multicolumn{4}{c}{D4RL}                & \multicolumn{2}{c}{Robomimic} \\
                                      & Locomotion & Kitchen & Adroit & AntMaze & MG          & PH, MH          \\ \midrule
    $\alpha$ for $Q^\beta$ + $s_\psi$ & 1          & 1       & 10     & 10      & 1           & 0.1             \\
    $\alpha$ for ARQ + $s_\psi$       & 1          & 1       & 10     & 10      & 1           & 0.1             \\ \midrule
    $\pi_\phi$ &
      \multicolumn{3}{c}{\begin{tabular}[c]{@{}c@{}}State-independent \\ stochastic \cite{IQL:kostrikov2021}\end{tabular}} &
      \multicolumn{1}{c}{Det.} &
      \multicolumn{2}{c}{Det.-ResNet} \\
    Training Iterations           & \multicolumn{4}{c}{1,000,000}           & \multicolumn{2}{c}{300,000} \\
    $\alpha$ for ARQ + $\pi_\phi$ & 1          & 1      & 10      & 10      & 10           & 0.1             \\
    \bottomrule
    \end{tabular}
\end{table}

\subsection{Full Experiment Results}

We display the full experiment results in Table~\ref{tab:full}.

\begin{table}[h]
    \centering
    \caption{Performance of prior methods and ours on D4RL\cite{D4RL} and Robomimic\cite{robomimic:Mandlekar:CoRL2021} datasets. Each number represents the performance relative to a random policy as $0$ and an expert policy as $100$. Unless noted as (ours) or (repro.), all the numbers are borrowed from \cite{IQL:kostrikov2021}, \cite{TD3BC:fujimoto:neurips2021}, and \cite{robomimic:Mandlekar:CoRL2021}. The numbers generated by us are averaged over 3 different random seeds. The standard deviations of multiple runs are also displayed.}
    \label{tab:full}
    \resizebox{\textwidth}{!}{%
    \begin{tabular}{@{}llcccccccccccccccc@{}}
    \toprule
     &  & \multicolumn{3}{c}{Without reward} & {\color[HTML]{9B9B9B} } & \multicolumn{2}{c}{With $Q^\beta$} & {\color[HTML]{9B9B9B} } & \multicolumn{9}{c}{With reward / value function} \\ \midrule
     &  & BC & \begin{tabular}[c]{@{}c@{}}Impl.\\ BC\end{tabular} & \begin{tabular}[c]{@{}c@{}}$s_\psi$\\ (ours)\end{tabular} & {\color[HTML]{9B9B9B} $\sigma$} & \begin{tabular}[c]{@{}c@{}}One-\\ step\end{tabular} & \begin{tabular}[c]{@{}c@{}}Q($\beta$)\\ +$s_\psi$\\ (ours)\end{tabular} & {\color[HTML]{9B9B9B} $\sigma$} & DT & AWAC & \begin{tabular}[c]{@{}c@{}}TD3\\ +BC\end{tabular} & CQL & IQL & \begin{tabular}[c]{@{}c@{}}ARQ\\ +$\pi_\phi$\\ (ours)\end{tabular} & {\color[HTML]{9B9B9B} $\sigma$} & \begin{tabular}[c]{@{}c@{}}ARQ\\ +$s_\psi$\\ (ours)\end{tabular} & {\color[HTML]{9B9B9B} $\sigma$} \\ \midrule
    expert-v2 & hopper & 112 & 110 & 89 & {\color[HTML]{9B9B9B} (5.1)} &  & 99 & {\color[HTML]{9B9B9B} (2.1)} &  & 85 & 108 & 111 & 110 & 111 & {\color[HTML]{9B9B9B} (0.2)} & 98 & {\color[HTML]{9B9B9B} (0.8)} \\
    & walker & 56 & 82 & 106 & {\color[HTML]{9B9B9B} (0.6)} &  & 107 & {\color[HTML]{9B9B9B} (0.4)} &  & 57 & 110 & 104 & 110 & 109 & {\color[HTML]{9B9B9B} (0.4)} & 108 & {\color[HTML]{9B9B9B} (0.1)} \\
    & halfchtah & 105 & 78 & 81 & {\color[HTML]{9B9B9B} (0.2)} &  & 84 & {\color[HTML]{9B9B9B} (1.1)} &  & 79 & 97 & 82 & 95 & 94 & {\color[HTML]{9B9B9B} (0.2)} & 85 & {\color[HTML]{9B9B9B} (0.5)} \\
    medium-v2 & hopper & 53 & 75 & 38 & {\color[HTML]{9B9B9B} (1.3)} & 60 & 55 & {\color[HTML]{9B9B9B} (0.6)} & 68 & 57 & 59 & 59 & 66 & 61 & {\color[HTML]{9B9B9B} (0.4)} & 58 & {\color[HTML]{9B9B9B} (0.8)} \\
    & walker & 75 & 15 & 63 & {\color[HTML]{9B9B9B} (1.6)} & 82 & 79 & {\color[HTML]{9B9B9B} (0.6)} & 74 & 72 & 84 & 73 & 78 & 81 & {\color[HTML]{9B9B9B} (0.7)} & 81 & {\color[HTML]{9B9B9B} (0.2)} \\
    & halfchtah & 43 & 35 & 40 & {\color[HTML]{9B9B9B} (0.3)} & 48 & 43 & {\color[HTML]{9B9B9B} (0.0)} & 43 & 44 & 48 & 44 & 47 & 45 & {\color[HTML]{9B9B9B} (0.3)} & 45 & {\color[HTML]{9B9B9B} (0.2)} \\
    med.-exp.-v2 & hopper & 53 & 28 & 45 & {\color[HTML]{9B9B9B} (0.4)} & 103 & 88 & {\color[HTML]{9B9B9B} (4.1)} & 108 & 56 & 98 & 105 & 92 & 110 & {\color[HTML]{9B9B9B} (0.9)} & 93 & {\color[HTML]{9B9B9B} (2.5)} \\
    & walker & 108 & 16 & 74 & {\color[HTML]{9B9B9B} (2.8)} & 113 & 107 & {\color[HTML]{9B9B9B} (0.2)} & 108 & 75 & 110 & 109 & 110 & 109 & {\color[HTML]{9B9B9B} (0.5)} & 107 & {\color[HTML]{9B9B9B} (0.4)} \\
    & halfchtah & 55 & 35 & 46 & {\color[HTML]{9B9B9B} (0.4)} & 93 & 77 & {\color[HTML]{9B9B9B} (1.8)} & 87 & 43 & 91 & 92 & 87 & 91 & {\color[HTML]{9B9B9B} (0.7)} & 82 & {\color[HTML]{9B9B9B} (0.3)} \\
    med.-rep.-v2 & hopper & 18 & 14 & 11 & {\color[HTML]{9B9B9B} (0.3)} & 98 & 62 & {\color[HTML]{9B9B9B} (0.9)} & 83 & 37 & 61 & 95 & 95 & 81 & {\color[HTML]{9B9B9B} (24.2)} & 78 & {\color[HTML]{9B9B9B} (3.2)} \\
    & walker & 26 & 10 & 20 & {\color[HTML]{9B9B9B} (0.3)} & 50 & 69 & {\color[HTML]{9B9B9B} (2.0)} & 67 & 27 & 82 & 77 & 74 & 66 & {\color[HTML]{9B9B9B} (7.0)} & 71 & {\color[HTML]{9B9B9B} (1.2)} \\
    & halfchtah & 37 & 25 & 26 & {\color[HTML]{9B9B9B} (0.8)} & 38 & 41 & {\color[HTML]{9B9B9B} (0.3)} & 37 & 41 & 45 & 46 & 44 & 42 & {\color[HTML]{9B9B9B} (0.3)} & 41 & {\color[HTML]{9B9B9B} (0.1)} \\ \midrule
    human-v0 & pen & 64 & 84 & 76 & {\color[HTML]{9B9B9B} (6.0)} &  & 73 & {\color[HTML]{9B9B9B} (1.9)} &  &  &  & 38 & 72 & 44 & {\color[HTML]{9B9B9B} (5.2)} & 74 & {\color[HTML]{9B9B9B} (1.3)} \\
    & door & 2 & 14 & 9 & {\color[HTML]{9B9B9B} (0.9)} &  & 11 & {\color[HTML]{9B9B9B} (1.4)} &  &  &  & 10 & 4 & 0 & {\color[HTML]{9B9B9B} (0.4)} & 10 & {\color[HTML]{9B9B9B} (4.9)} \\
    & relocate & 0 & 0 & 0 & {\color[HTML]{9B9B9B} (0.0)} &  & 0 & {\color[HTML]{9B9B9B} (0.0)} &  &  &  & 0 & 0 & 0 & {\color[HTML]{9B9B9B} (0.0)} & 0 & {\color[HTML]{9B9B9B} (0.1)} \\
    & hammer & 1 & 1 & 2 & {\color[HTML]{9B9B9B} (0.4)} &  & 5 & {\color[HTML]{9B9B9B} (1.2)} &  &  &  & 4 & 1 & 1 & {\color[HTML]{9B9B9B} (0.7)} & 6 & {\color[HTML]{9B9B9B} (1.0)} \\
    cloned-v0 & pen & 37 &  & 26 & {\color[HTML]{9B9B9B} (4.3)} & 60 & 57 & {\color[HTML]{9B9B9B} (2.3)} &  &  &  & 39 & 37 & 48 & {\color[HTML]{9B9B9B} (7.1)} & 55 & {\color[HTML]{9B9B9B} (2.7)} \\
    & door & 0 &  & 2 & {\color[HTML]{9B9B9B} (0.4)} & 0 & 11 & {\color[HTML]{9B9B9B} (2.0)} &  &  &  & 0 & 2 & 1 & {\color[HTML]{9B9B9B} (0.4)} & 12 & {\color[HTML]{9B9B9B} (4.4)} \\
    & relocate & 0 &  & 0 & {\color[HTML]{9B9B9B} (0.0)} & 0 & 0 & {\color[HTML]{9B9B9B} (0.0)} &  &  &  & 0 & 0 & 0 & {\color[HTML]{9B9B9B} (0.0)} & 0 & {\color[HTML]{9B9B9B} (0.0)} \\
    & hammer & 1 &  & 1 & {\color[HTML]{9B9B9B} (0.2)} & 2 & 3 & {\color[HTML]{9B9B9B} (0.3)} &  &  &  & 2 & 2 & 1 & {\color[HTML]{9B9B9B} (0.3)} & 4 & {\color[HTML]{9B9B9B} (2.6)} \\ \midrule
    kitchen-v0 & complete & 65 & 85 & 74 & {\color[HTML]{9B9B9B} (3.7)} &  & 75 & {\color[HTML]{9B9B9B} (1.2)} &  &  &  & 44 & 63 & 37 & {\color[HTML]{9B9B9B} (14.2)} & 77 & {\color[HTML]{9B9B9B} (1.8)} \\
    & partial & 38 & 38 & 45 & {\color[HTML]{9B9B9B} (2.7)} &  & 59 & {\color[HTML]{9B9B9B} (4.9)} &  &  &  & 50 & 46 & 50 & {\color[HTML]{9B9B9B} (5.0)} & 70 & {\color[HTML]{9B9B9B} (2.7)} \\
    & mixed & 52 & 38 & 51 & {\color[HTML]{9B9B9B} (1.1)} &  & 52 & {\color[HTML]{9B9B9B} (1.1)} &  &  &  & 51 & 51 & 39 & {\color[HTML]{9B9B9B} (9.4)} & 57 & {\color[HTML]{9B9B9B} (2.9)} \\ \midrule
    antmaze-v0 & umaze & 55 &  & 58 & {\color[HTML]{9B9B9B} (2.1)} & 64 & 81 & {\color[HTML]{9B9B9B} (4.5)} & 59 & 57 & 79 & 74 & 88 & 97 & {\color[HTML]{9B9B9B} (0.8)} & 94 & {\color[HTML]{9B9B9B} (1.7)} \\
    & umaze-div. & 46 &  & 61 & {\color[HTML]{9B9B9B} (1.4)} & 61 & 62 & {\color[HTML]{9B9B9B} (3.3)} & 53 & 49 & 71 & 84 & 62 & 62 & {\color[HTML]{9B9B9B} (12.1)} & 58 & {\color[HTML]{9B9B9B} (7.0)} \\
    & med.-play & 0 &  & 1 & {\color[HTML]{9B9B9B} (0.5)} & 0 & 25 & {\color[HTML]{9B9B9B} (13.3)} & 0 & 0 & 11 & 61 & 71 & 80 & {\color[HTML]{9B9B9B} (8.3)} & 69 & {\color[HTML]{9B9B9B} (6.6)} \\
    & med.-div. & 0 &  & 1 & {\color[HTML]{9B9B9B} (0.5)} & 0 & 45 & {\color[HTML]{9B9B9B} (5.3)} & 0 & 1 & 3 & 54 & 70 & 82 & {\color[HTML]{9B9B9B} (6.1)} & 65 & {\color[HTML]{9B9B9B} (15.6)} \\
    & large-play & 0 &  & 0 & {\color[HTML]{9B9B9B} (0.0)} & 0 & 1 & {\color[HTML]{9B9B9B} (0.5)} & 0 & 0 & 0 & 16 & 40 & 37 & {\color[HTML]{9B9B9B} (17.7)} & 18 & {\color[HTML]{9B9B9B} (1.7)} \\
    & large-div. & 0 &  & 0 & {\color[HTML]{9B9B9B} (0.0)} & 0 & 1 & {\color[HTML]{9B9B9B} (0.5)} & 0 & 1 & 0 & 15 & 48 & 58 & {\color[HTML]{9B9B9B} (6.2)} & 23 & {\color[HTML]{9B9B9B} (5.0)} \\ \toprule
     &  & BC$^\star$ &  & \begin{tabular}[c]{@{}c@{}}$s_\psi$\\ (ours)\end{tabular} & {\color[HTML]{9B9B9B} } &  & \begin{tabular}[c]{@{}c@{}}Q($\beta$)\\ +$s_\psi$\\ (ours)\end{tabular} & {\color[HTML]{9B9B9B} } &  &  & BCQ$^\star$ & CQL$^\star$ & \begin{tabular}[c]{@{}c@{}}IQL\\ (repro.)\end{tabular} & \begin{tabular}[c]{@{}c@{}}ARQ\\ +$\pi_\phi$\\ (ours)\end{tabular} & {\color[HTML]{9B9B9B} } & \begin{tabular}[c]{@{}c@{}}ARQ\\ +$s_\psi$\\ (ours)\end{tabular} & {\color[HTML]{9B9B9B} } \\ \midrule
    machine gen. & lift & 65 & \multicolumn{1}{l}{} & 29 & {\color[HTML]{9B9B9B} (2.4)} & \multicolumn{1}{l}{} & 86 & {\color[HTML]{9B9B9B} (0.5)} & \multicolumn{1}{l}{} & \multicolumn{1}{l}{} & 91 & 64 & 79 & 79 & {\color[HTML]{9B9B9B} (1.2)} & 82 & {\color[HTML]{9B9B9B} (0.5)} \\
    & can & 65 & \multicolumn{1}{l}{} & 19 & {\color[HTML]{9B9B9B} (2.4)} & \multicolumn{1}{l}{} & 55 & {\color[HTML]{9B9B9B} (6.5)} & \multicolumn{1}{l}{} & \multicolumn{1}{l}{} & 75 & 1 & 62 & 76 & {\color[HTML]{9B9B9B} (0.5)} & 60 & {\color[HTML]{9B9B9B} (1.2)} \\ \midrule
    pro. human & lift & 100 & \multicolumn{1}{l}{} & 99 & {\color[HTML]{9B9B9B} (0.8)} & \multicolumn{1}{l}{} & 100 & {\color[HTML]{9B9B9B} (0.5)} & \multicolumn{1}{l}{} & \multicolumn{1}{l}{} & 100 & 93 & 58 & 100 & {\color[HTML]{9B9B9B} (0.0)} & 98 & {\color[HTML]{9B9B9B} (0.0)} \\
    & can & 95 & \multicolumn{1}{l}{} & 95 & {\color[HTML]{9B9B9B} (0.8)} & \multicolumn{1}{l}{} & 93 & {\color[HTML]{9B9B9B} (2.4)} & \multicolumn{1}{l}{} & \multicolumn{1}{l}{} & 89 & 38 & 26 & 92 & {\color[HTML]{9B9B9B} (2.2)} & 95 & {\color[HTML]{9B9B9B} (0.8)} \\
    & square & 79 & \multicolumn{1}{l}{} & 66 & {\color[HTML]{9B9B9B} (2.9)} & \multicolumn{1}{l}{} & 72 & {\color[HTML]{9B9B9B} (4.0)} & \multicolumn{1}{l}{} & \multicolumn{1}{l}{} & 50 & 5 & 24 & 44 & {\color[HTML]{9B9B9B} (2.2)} & 69 & {\color[HTML]{9B9B9B} (3.4)} \\
    & transport & 17 & \multicolumn{1}{l}{} & 27 & {\color[HTML]{9B9B9B} (2.5)} & \multicolumn{1}{l}{} & 28 & {\color[HTML]{9B9B9B} (8.7)} & \multicolumn{1}{l}{} & \multicolumn{1}{l}{} & 7 & 0 & 1 & 29 & {\color[HTML]{9B9B9B} (5.0)} & 30 & {\color[HTML]{9B9B9B} (5.4)} \\
    & tool-hang & 29 & \multicolumn{1}{l}{} & 70 & {\color[HTML]{9B9B9B} (4.2)} & \multicolumn{1}{l}{} & 64 & {\color[HTML]{9B9B9B} (2.9)} & \multicolumn{1}{l}{} & \multicolumn{1}{l}{} & 0 & 0 & 3 & 3 & {\color[HTML]{9B9B9B} (1.7)} & 71 & {\color[HTML]{9B9B9B} (5.9)} \\ \midrule
    multi. human & lift & 100 & \multicolumn{1}{l}{} & 96 & {\color[HTML]{9B9B9B} (1.7)} & \multicolumn{1}{l}{} & 94 & {\color[HTML]{9B9B9B} (3.6)} & \multicolumn{1}{l}{} & \multicolumn{1}{l}{} & 100 & 57 & 51 & 99 & {\color[HTML]{9B9B9B} (1.4)} & 95 & {\color[HTML]{9B9B9B} (2.5)} \\
    & can & 86 & \multicolumn{1}{l}{} & 84 & {\color[HTML]{9B9B9B} (1.2)} & \multicolumn{1}{l}{} & 89 & {\color[HTML]{9B9B9B} (1.6)} & \multicolumn{1}{l}{} & \multicolumn{1}{l}{} & 63 & 22 & 25 & 90 & {\color[HTML]{9B9B9B} (1.2)} & 86 & {\color[HTML]{9B9B9B} (1.7)} \\
    & square & 53 & \multicolumn{1}{l}{} & 44 & {\color[HTML]{9B9B9B} (1.7)} & \multicolumn{1}{l}{} & 51 & {\color[HTML]{9B9B9B} (4.9)} & \multicolumn{1}{l}{} & \multicolumn{1}{l}{} & 14 & 1 & 12 & 31 & {\color[HTML]{9B9B9B} (4.2)} & 51 & {\color[HTML]{9B9B9B} (4.8)} \\
    & transport & 11 & \multicolumn{1}{l}{} & 15 & {\color[HTML]{9B9B9B} (3.1)} & \multicolumn{1}{l}{} & 17 & {\color[HTML]{9B9B9B} (2.6)} & \multicolumn{1}{l}{} & \multicolumn{1}{l}{} & 3 & 0 & 0 & 16 & {\color[HTML]{9B9B9B} (4.9)} & 13 & {\color[HTML]{9B9B9B} (1.7)} \\ \midrule
    \multicolumn{18}{r}{$\star$ represents that the best performance during training iterations is picked a posteriori.}
    \end{tabular}%
    }
    \end{table}

%% file: root.bbl
\begin{thebibliography}{10}

\bibitem{onestep:brandfonbrener:neurips2021}
David Brandfonbrener, William~F Whitney, Rajesh Ranganath, and Joan Bruna.
\newblock Offline {RL} without off-policy evaluation.
\newblock In A.~Beygelzimer, Y.~Dauphin, P.~Liang, and J.~Wortman Vaughan,
  editors, {\em Advances in Neural Information Processing Systems}, 2021.

\bibitem{openaigym}
Greg Brockman, Vicki Cheung, Ludwig Pettersson, Jonas Schneider, John Schulman,
  Jie Tang, and Wojciech Zaremba.
\newblock Openai gym.
\newblock {\em CoRR}, abs/1606.01540, 2016.

\bibitem{brown2021value}
Daniel~S Brown, Jordan Schneider, Anca Dragan, and Scott Niekum.
\newblock Value alignment verification.
\newblock In {\em Proceedings of the 38th International Conference on Machine
  Learning}, pages 1105--1115, 2021.

\bibitem{pessimism}
Jacob Buckman, Carles Gelada, and Marc~G Bellemare.
\newblock The importance of pessimism in fixed-dataset policy optimization.
\newblock {\em arXiv preprint arXiv:2009.06799}, 2020.

\bibitem{DecisionTransformer:chen:neurips2021}
Lili Chen, Kevin Lu, Aravind Rajeswaran, Kimin Lee, Aditya Grover, Misha
  Laskin, Pieter Abbeel, Aravind Srinivas, and Igor Mordatch.
\newblock Decision transformer: Reinforcement learning via sequence modeling.
\newblock {\em Advances in neural information processing systems}, 34, 2021.

\bibitem{PLOFF:Dadashi:ICML21}
Robert Dadashi, Shideh Rezaeifar, Nino Vieillard, Léonard Hussenot, Olivier
  Pietquin, and Matthieu Geist.
\newblock Offline reinforcement learning with pseudometric learning.
\newblock In {\em ICML}, pages 2307--2318, 2021.

\bibitem{ding2019goal}
Yiming Ding, Carlos Florensa, Pieter Abbeel, and Mariano Phielipp.
\newblock Goal-conditioned imitation learning.
\newblock {\em Advances in neural information processing systems}, 32, 2019.

\bibitem{rvs:emmons:ICLR2022}
Scott Emmons, Benjamin Eysenbach, Ilya Kostrikov, and Sergey Levine.
\newblock Rvs: What is essential for offline {RL} via supervised learning?
\newblock In {\em International Conference on Learning Representations}, 2022.

\bibitem{implicitBC:florence:CoRL2021}
Pete Florence, Corey Lynch, Andy Zeng, Oscar~A Ramirez, Ayzaan Wahid, Laura
  Downs, Adrian Wong, Johnny Lee, Igor Mordatch, and Jonathan Tompson.
\newblock Implicit behavioral cloning.
\newblock In {\em 5th Annual Conference on Robot Learning}, 2021.

\bibitem{D4RL}
Justin Fu, Aviral Kumar, Ofir Nachum, George Tucker, and Sergey Levine.
\newblock D4rl: Datasets for deep data-driven reinforcement learning.
\newblock {\em arXiv preprint arXiv:2004.07219}, 2020.

\bibitem{TD3BC:fujimoto:neurips2021}
Scott Fujimoto and Shixiang~Shane Gu.
\newblock A minimalist approach to offline reinforcement learning.
\newblock In {\em Thirty-Fifth Conference on Neural Information Processing
  Systems}, 2021.

\bibitem{BCQ}
Scott Fujimoto, David Meger, and Doina Precup.
\newblock Off-policy deep reinforcement learning without exploration.
\newblock In {\em International Conference on Machine Learning}, pages
  2052--2062, 2019.

\bibitem{ghosh2019learning}
Dibya Ghosh, Abhishek Gupta, Ashwin Reddy, Justin Fu, Coline Devin, Benjamin
  Eysenbach, and Sergey Levine.
\newblock Learning to reach goals via iterated supervised learning.
\newblock {\em arXiv preprint arXiv:1912.06088}, 2019.

\bibitem{Goo2021YOEO}
Wonjoon Goo and Scott Niekum.
\newblock You only evaluate once: a simple baseline algorithm for offline {RL}.
\newblock In {\em 5th Annual Conference on Robot Learning}, 2021.

\bibitem{goodfellow2014generative}
Ian Goodfellow, Jean Pouget-Abadie, Mehdi Mirza, Bing Xu, David Warde-Farley,
  Sherjil Ozair, Aaron Courville, and Yoshua Bengio.
\newblock Generative adversarial nets.
\newblock {\em Advances in neural information processing systems}, 27, 2014.

\bibitem{frankakitchen}
Abhishek Gupta, Vikash Kumar, Corey Lynch, Sergey Levine, and Karol Hausman.
\newblock Relay policy learning: Solving long horizon tasks via imitation and
  reinforcement learning.
\newblock {\em Conference on Robot Learning (CoRL)}, 2019.

\bibitem{SAC}
Tuomas Haarnoja, Aurick Zhou, Pieter Abbeel, and Sergey Levine.
\newblock Soft actor-critic: Off-policy maximum entropy deep reinforcement
  learning with a stochastic actor.
\newblock In {\em International Conference on Machine Learning}, pages
  1861--1870, 2018.

\bibitem{norlnosim:hahn:neurips2021}
Meera Hahn, Devendra~Singh Chaplot, Shubham Tulsiani, Mustafa Mukadam,
  James~Matthew Rehg, and Abhinav Gupta.
\newblock No {RL}, no simulation: Learning to navigate without navigating.
\newblock In A.~Beygelzimer, Y.~Dauphin, P.~Liang, and J.~Wortman Vaughan,
  editors, {\em Advances in Neural Information Processing Systems}, 2021.

\bibitem{he2015deep}
Kaiming He, Xiangyu Zhang, Shaoqing Ren, and Jian Sun.
\newblock Deep residual learning for image recognition, 2015.

\bibitem{hyvarinen2005estimation}
Aapo Hyv{\"a}rinen and Peter Dayan.
\newblock Estimation of non-normalized statistical models by score matching.
\newblock {\em Journal of Machine Learning Research}, 6(4), 2005.

\bibitem{morel}
Rahul Kidambi, Aravind Rajeswaran, Praneeth Netrapalli, and Thorsten Joachims.
\newblock Morel: Model-based offline reinforcement learning.
\newblock In H.~Larochelle, M.~Ranzato, R.~Hadsell, M.~F. Balcan, and H.~Lin,
  editors, {\em Advances in Neural Information Processing Systems}, volume~33,
  pages 21810--21823. Curran Associates, Inc., 2020.

\bibitem{vae:kingma2013}
Diederik~P Kingma and Max Welling.
\newblock Auto-encoding variational bayes.
\newblock {\em arXiv preprint arXiv:1312.6114}, 2013.

\bibitem{fishercql:kostrikov:icml2021}
Ilya Kostrikov, Rob Fergus, Jonathan Tompson, and Ofir Nachum.
\newblock Offline reinforcement learning with fisher divergence critic
  regularization.
\newblock In {\em International Conference on Machine Learning}, pages
  5774--5783. PMLR, 2021.

\bibitem{IQL:kostrikov2021}
Ilya Kostrikov, Ashvin Nair, and Sergey Levine.
\newblock Offline reinforcement learning with implicit q-learning.
\newblock {\em arXiv preprint arXiv:2110.06169}, 2021.

\bibitem{BEAR}
Aviral Kumar, Justin Fu, Matthew Soh, George Tucker, and Sergey Levine.
\newblock Stabilizing off-policy q-learning via bootstrapping error reduction.
\newblock In {\em Advances in Neural Information Processing Systems 32}, pages
  11784--11794, 2019.

\bibitem{whenshould:kumar:ICLR2022}
Aviral Kumar, Joey Hong, Anikait Singh, and Sergey Levine.
\newblock Should i run offline reinforcement learning or behavioral cloning?
\newblock In {\em International Conference on Learning Representations}, 2022.

\bibitem{kumar2019reward}
Aviral Kumar, Xue~Bin Peng, and Sergey Levine.
\newblock Reward-conditioned policies.
\newblock {\em arXiv preprint arXiv:1912.13465}, 2019.

\bibitem{cql-workflow:kumar:corl2021}
Aviral Kumar, Anikait Singh, Stephen Tian, Chelsea Finn, and Sergey Levine.
\newblock A workflow for offline model-free robotic reinforcement learning.
\newblock In {\em 5th Annual Conference on Robot Learning}, 2021.

\bibitem{CQL}
Aviral Kumar, Aurick Zhou, George Tucker, and Sergey Levine.
\newblock Conservative q-learning for offline reinforcement learning.
\newblock {\em arXiv preprint arXiv:2006.04779}, 2020.

\bibitem{lagoudakis2003least}
Michail~G Lagoudakis and Ronald Parr.
\newblock Least-squares policy iteration.
\newblock {\em The Journal of Machine Learning Research}, 4:1107--1149, 2003.

\bibitem{lecun2006tutorial}
Yann LeCun, Sumit Chopra, Raia Hadsell, M~Ranzato, and F~Huang.
\newblock A tutorial on energy-based learning.
\newblock {\em Predicting structured data}, 1(0), 2006.

\bibitem{levine2020offRLtuto}
Sergey Levine, Aviral Kumar, George Tucker, and Justin Fu.
\newblock Offline reinforcement learning: Tutorial, review, and perspectives on
  open problems.
\newblock {\em arXiv preprint arXiv:2005.01643}, 2020.

\bibitem{liu2016kernelized}
Qiang Liu, Jason Lee, and Michael Jordan.
\newblock A kernelized stein discrepancy for goodness-of-fit tests.
\newblock In {\em International conference on machine learning}, pages
  276--284. PMLR, 2016.

\bibitem{MBS:Liu:neurips2020}
Yao Liu, Adith Swaminathan, Alekh Agarwal, and Emma Brunskill.
\newblock Provably good batch reinforcement learning without great exploration.
\newblock In {\em Proceedings of the 34th International Conference on Neural
  Information Processing Systems}, 2020.

\bibitem{robomimic:Mandlekar:CoRL2021}
Ajay Mandlekar, Danfei Xu, Josiah Wong, Soroush Nasiriany, Chen Wang, Rohun
  Kulkarni, Li~Fei-Fei, Silvio Savarese, Yuke Zhu, and Roberto
  Mart{\'\i}n-Mart{\'\i}n.
\newblock What matters in learning from offline human demonstrations for robot
  manipulation.
\newblock In {\em 5th Annual Conference on Robot Learning}, 2021.

\bibitem{AWR}
Xue~Bin Peng, Aviral Kumar, Grace Zhang, and Sergey Levine.
\newblock Advantage-weighted regression: Simple and scalable off-policy
  reinforcement learning.
\newblock {\em arXiv preprint arXiv:1910.00177}, 2019.

\bibitem{adroit}
Aravind Rajeswaran, Vikash Kumar, Abhishek Gupta, Giulia Vezzani, John
  Schulman, Emanuel Todorov, and Sergey Levine.
\newblock {Learning Complex Dexterous Manipulation with Deep Reinforcement
  Learning and Demonstrations}.
\newblock In {\em Proceedings of Robotics: Science and Systems (RSS)}, 2018.

\bibitem{bridging:rashidinejad:neurips2021}
Paria Rashidinejad, Banghua Zhu, Cong Ma, Jiantao Jiao, and Stuart Russell.
\newblock Bridging offline reinforcement learning and imitation learning: A
  tale of pessimism.
\newblock {\em Advances in Neural Information Processing Systems}, 34, 2021.

\bibitem{anti-exploration:rezaeifar2021offline}
Shideh Rezaeifar, Robert Dadashi, Nino Vieillard, Léonard Hussenot, Olivier
  Bachem, Olivier Pietquin, and Matthieu Geist.
\newblock Offline reinforcement learning as anti-exploration, 2021.

\bibitem{schaal1999imitation}
Stefan Schaal.
\newblock Is imitation learning the route to humanoid robots?
\newblock {\em Trends in cognitive sciences}, 3(6):233--242, 1999.

\bibitem{trpo:schulman2015trust}
John Schulman, Sergey Levine, Pieter Abbeel, Michael Jordan, and Philipp
  Moritz.
\newblock Trust region policy optimization.
\newblock In {\em International conference on machine learning}, pages
  1889--1897. PMLR, 2015.

\bibitem{ppo:schulman2017proximal}
John Schulman, Filip Wolski, Prafulla Dhariwal, Alec Radford, and Oleg Klimov.
\newblock Proximal policy optimization algorithms.
\newblock {\em arXiv preprint arXiv:1707.06347}, 2017.

\bibitem{score:song:neurips2019}
Yang Song and Stefano Ermon.
\newblock Generative modeling by estimating gradients of the data distribution.
\newblock In {\em Advances in Neural Information Processing Systems}, pages
  11895--11907, 2019.

\bibitem{scorev2:song:neurips2020}
Yang Song and Stefano Ermon.
\newblock Improved techniques for training score-based generative models.
\newblock In Hugo Larochelle, Marc'Aurelio Ranzato, Raia Hadsell,
  Maria{-}Florina Balcan, and Hsuan{-}Tien Lin, editors, {\em Advances in
  Neural Information Processing Systems 33: Annual Conference on Neural
  Information Processing Systems 2020, NeurIPS 2020, December 6-12, 2020,
  virtual}, 2020.

\bibitem{sde:song:iclr2021}
Yang Song, Jascha Sohl-Dickstein, Diederik~P Kingma, Abhishek Kumar, Stefano
  Ermon, and Ben Poole.
\newblock Score-based generative modeling through stochastic differential
  equations.
\newblock In {\em International Conference on Learning Representations}, 2021.

\bibitem{srivastava2019training}
Rupesh~Kumar Srivastava, Pranav Shyam, Filipe Mutz, Wojciech Ja{\'s}kowski, and
  J{\"u}rgen Schmidhuber.
\newblock Training agents using upside-down reinforcement learning.
\newblock {\em arXiv preprint arXiv:1912.02877}, 2019.

\bibitem{suttonbartoRL}
Richard~S. Sutton and Andrew~G. Barto.
\newblock {\em Reinforcement Learning: An Introduction}.
\newblock MIT Press, Cambridge, MA, USA, 2018.

\bibitem{vincent2011connection}
Pascal Vincent.
\newblock A connection between score matching and denoising autoencoders.
\newblock {\em Neural computation}, 23(7):1661--1674, 2011.

\bibitem{CRR:Wang:2020}
Ziyu Wang, Alexander Novikov, Konrad Zolna, Josh~S Merel, Jost~Tobias
  Springenberg, Scott~E Reed, Bobak Shahriari, Noah Siegel, Caglar Gulcehre,
  Nicolas Heess, and Nando de~Freitas.
\newblock Critic regularized regression.
\newblock In H.~Larochelle, M.~Ranzato, R.~Hadsell, M.~F. Balcan, and H.~Lin,
  editors, {\em Advances in Neural Information Processing Systems}, volume~33,
  pages 7768--7778. Curran Associates, Inc., 2020.

\bibitem{policygradient-williams1992}
Ronald~J Williams.
\newblock Simple statistical gradient-following algorithms for connectionist
  reinforcement learning.
\newblock {\em Machine learning}, 8(3):229--256, 1992.

\bibitem{BRAC}
Yifan Wu, George Tucker, and Ofir Nachum.
\newblock Behavior regularized offline reinforcement learning.
\newblock {\em arXiv preprint arXiv:1911.11361}, 2019.

\bibitem{MOPO}
Tianhe Yu, Garrett Thomas, Lantao Yu, Stefano Ermon, James Zou, Sergey Levine,
  Chelsea Finn, and Tengyu Ma.
\newblock Mopo: Model-based offline policy optimization.
\newblock {\em arXiv preprint arXiv:2005.13239}, 2020.

\end{thebibliography}
